\documentclass{article}

\usepackage{arxiv}

\usepackage[utf8]{inputenc} 
\usepackage[T1]{fontenc}    
\usepackage{lmodern}        
\usepackage{hyperref}       
\usepackage{url}            
\usepackage{booktabs}       
\usepackage{amsfonts}       
\usepackage{nicefrac}       
\usepackage{microtype}      
\usepackage{graphicx}

\title{Opening the random forest black box by the analysis of the mutual
impact of features\,\,\,\,\,\,\,}

\author{
    Lucas F. Voges
   \\
    Centre for the Study of Manuscript Cultures (CSMC) \\
    Universität Hamburg \\
  Hamburg, Germany \\
  \texttt{} \\
   \And
    Lukas C. Jarren
   \\
    Centre for the Study of Manuscript Cultures (CSMC) / Hamburg School
of Food Science \\
    Universität Hamburg \\
  Hamburg, Germany \\
  \texttt{} \\
   \And
    Stephan Seifert
   \\
    Centre for the Study of Manuscript Cultures (CSMC) / Hamburg School
of Food Science \\
    Universität Hamburg \\
  Hamburg, Germany \\
  \texttt{\href{mailto:stephan.seifert@uni-hamburg.de}{\nolinkurl{stephan.seifert@uni-hamburg.de}}} \\
  }

\newlength{\cslhangindent}
\newlength{\csllabelwidth}
\newlength{\cslentryspacingunit} 
  {}%
  {\par}
\newenvironment{CSLReferences}[2] 
 {
  \setlength{\parindent}{0pt}
  \ifodd #1
  \let\oldpar\par
  \def\par{\hangindent=\cslhangindent\oldpar}
  \fi
  \setlength{\parskip}{#2\cslentryspacingunit}
 }%
 {}
\usepackage{calc}

\begin{document}
\maketitle

\begin{abstract}
Random forest is a popular machine learning approach for the analysis of
high-dimensional data because it is flexible and provides variable
importance measures for the selection of relevant features. However, the
complex relationships between the features are usually not considered
for the selection and thus also neglected for the characterization of
the analysed samples. Here we propose two novel approaches that focus on
the mutual impact of features in random forests. Mutual forest impact
(MFI) is a relation parameter that evaluates the mutual association of
the featurs to the outcome and, hence, goes beyond the analysis of
correlation coefficients. Mutual impurity reduction (MIR) is an
importance measure that combines this relation parameter with the
importance of the individual features. MIR and MFI are implemented
together with testing procedures that generate p-values for the
selection of related and important features. Applications to various
simulated data sets and the comparison to other methods for feature
selection and relation analysis show that MFI and MIR are very promising
to shed light on the complex relationships between features and outcome.
In addition, they are not affected by common biases, e.g.~that features
with many possible splits or high minor allele frequencies are prefered.
\newline \newline The approaches are implemented in Version 0.3.0 of the
R package RFSurrogates that is available at
github.com/StephanSeifert/RFSurrogates.
\end{abstract}

\newpage

\hypertarget{introduction}{%
\section{Introduction}\label{introduction}}

The application of the machine learning approach random forest (RF)
(Breiman 2001) has become very popular for the analysis of
high-dimensional data, e.g.~generated in genomics (X. Chen and Ishwaran
2012) and metabolomics (T. Chen et al. 2013) experiments or genome-wide
association studies (GWAS) (Nicholls et al. 2020). The reason for this
popularity are specific advantages over other methods, such as the
flexibility in terms of input and output variables, since both
quantitative and qualitative variables can be used to build
classification, regression (Strobl, Malley, and Tutz 2009) and survival
models (Ishwaran et al. 2011). Another advantage is the ability to
generate variable importance measures (VIMs) that are utilized to select
the relevant features for parsimonious prediction models or to identify
and interpret differences between the samples. \newline The most common
VIMs are the permutation importance and the impurity importance, so
called since it is calculated from the impurity gain that the variable
contributes to the random forest. Another importance measure is minimal
depth, which is based on the position of the variables in the decision
trees (Ishwaran et al. 2010). Minimal depth and the impurity importance,
however, are known to be biased in favour of features with many
categories (Strobl et al. 2007) and high category frequencies (K. K.
Nicodemus 2011), which is particularly important in GWAS (Boulesteix et
al. 2012). Since it is not affected by these biases, the permutation
importance has been preferred and various selection techniques based on
this importance measure have been developed (Szymczak et al. 2016;
Janitza, Celik, and Boulesteix 2018; Kursa and Rudnicki 2010) and
compared (Degenhardt, Seifert, and Szymczak 2019). A few years ago, a
corrected, unbiased impurity importance measure, the actual impurity
reduction (AIR), was introduced (Nembrini, König, and Wright 2018). AIR
is computed faster than the permutation importance, which is why this
importance measure is very useful for application to high-dimensional
data. \newline All of the RF based importance measures are affected by
the correlation structure of the features (Kristin K. Nicodemus et al.
2010), and conditional variable importance has been proposed to
determine the corrected, individual impact of the features (Strobl et
al. 2008; Debeer and Strobl 2020). We have taken a different approach to
this issue because we think that the relations between the features
should be included into the analysis treating them as interacting
components. Therefore, we have deliberately included feature relations
to improve variable importance calculation, the power of feature
selection and interpretation of differences between samples (Seifert,
Gundlach, and Szymczak 2019). We have achieved this by the exploitation
of surrogate variables that have been introduced to compensate for
missing values in the data, representing the features that can replace
another feature in a split as best as possible (Breiman et al. 1984).
\newline Based on surrogate variables, we developed Surrogate Minimal
Depth (SMD), an importance measure that incorporates surrogate variables
into the concept of minimal depth, and mean adjusted agreement, a
relation parameter that is determined by the split agreement of the
features across the random forest. Since this relation parameter
considers the mutual impact of the features on the random forest model,
this parameter goes beyond the analysis of ordinary correlation
coefficients enabling a comprehensive analysis of the complex interplay
of features and outcome. We applied this relation analysis to reveil
relations between features in gene expression (Seifert, Gundlach, and
Szymczak 2019), metabolomics (Wenck et al. 2022), and various
spectroscopic data sets (Shakiba et al. 2022; Seifert 2020), e.g.~to
illuminate the interaction of drugs with proteins and lipids in living
cells (Zivanovic et al. 2019). However, since both SMD and mean adjusted
agreement are affected by the previously described biases, their
application has so far been limited. Here we introduce two novel
approaches for the analysis of feature relations and mutual importance
called mutual forest impact (MFI) and mutual impurity reduction (MIR).
We will show that they are not affected by these biases and compare
their performance with existing approaches by applying them to different
simulated data sets.

\hypertarget{materials-and-methods}{%
\section{Materials and Methods}\label{materials-and-methods}}

\hypertarget{random-forest}{%
\subsection{Random Forest}\label{random-forest}}

RF is an ensemble of binary decision trees for classification,
regression and survival analysis (Breiman 2001). Each of these decision
trees is built from a different bootstrap sample and the best split for
each node is identified from randomly chosen candidate features by
maximizing the decrease of impurity. The impurity reduction is usually
determined by the Gini index for classification (Breiman et al. 1984),
the sum of squares for regression (Ishwaran 2015) and the log-rank
statistic for survival analysis (Ishwaran et al. 2008). The three
important parameters of RF are \emph{ntree}, the total number of
decision trees, \emph{mtry}, the number of randomly chosen candidate
features, and \emph{min.node.size}, the maximum number of samples of the
terminal nodes. Since RF is based on bagging, there are out-of-bag (OOB)
samples for each tree, which have not been used in the training process
and therefore can be utilized to determine prediction accuracy and
variable importance.

\hypertarget{impurity-importance}{%
\subsection{Impurity importance}\label{impurity-importance}}

The impurity importance is based on the decrease of impurity determined
by the difference of a node's impurity and the weighted sum of the child
node's impurities. To determine the importance of a feature, the sum of
all impurity decrease measures of the nodes based on this feature is
divided by the number of trees. The impurity importance is biased in
favour of features with many possible split points because they have a
higher probability to be randomly suitable for the intended distinction
(Strobl et al. 2007; Wright, Dankowski, and Ziegler 2017). In addition,
features with the same number of categories but different category
frequencies are biased as well (K. K. Nicodemus 2011), which is crucial
for genetic analyses because single-nucleotide polymorphisms (SNPs) with
high minor allele frequencies (MAF) are favoured (Boulesteix et al.
2012). In order to correct this bias, Nembrini, König, and Wright (2018)
applied a modified approach of Sandri and Zuccolotto (2008) to calculate
the AIR of each feature \(X_i\) by the difference between the VIM and
the VIM of a permuted version of itself:

\[\widehat{AIR_{X_{i}}} =  \widehat{VIM_{X_{i}}} - \widehat{VIM_{X_{i,perm}}}\]

\hypertarget{surrogate-variables-for-relation-analysis}{%
\subsection{Surrogate variables for relation
analysis}\label{surrogate-variables-for-relation-analysis}}

Surrogate variables were introduced by Breiman et al. (1984) for the
compensation of missing values. The basic idea is that in addition to
the primary splits of the decision trees in RF, alternative splits are
determined based on other features of the data that can replace the
primary split as best as possible. For the selection of the predefined
number \emph{s} of surrogate splits, the surrogates with the highest
values for the adjusted agreement \emph{adj} are chosen. This parameter
is calculated for the primary split \emph{p} and the possible surrogate
\emph{q} utilizing the agreement \emph{\(n_{surr}\)}, which is
determined by the number of samples that are assigned to the same
daughter nodes. It is defined by

\[adj(p,q) = \frac{n_{surr} - n_{maj}}{n_{total} - n_{maj}} \]

where \emph{\(n_{total}\)} is the total number of samples at the
respective node and \emph{\(n_{maj}\)} is the number of correct
assignments when all samples are assigned to the daughter node with the
larger number of samples, also called the majority rule. Note that
surrogate variables are only chosen when the adjusted agreement is
higher than 0 meaning that the surrogate split has to outperform the
majority rule. This can result in less than s surrogate splits for
individual nodes. \newline  For the analysis of the variable relations
of \(X_j\) to \(X_i\), all nodes with primary split variable \(X_i\) are
considered and the mean adjusted agreement \(M_{X_{i} X_{j}}\) is
defined by

\[
M_{X_{i} X_{j}} = \frac{\sum_{n=1}^{|nodes(X_{i} X_{j})|}agree(p_n^{X_{i}},q_n^{X_{j}})}{|nodes(X_{i})|}
\]

with \(p_n^{X_{i}}\) and \(q_n^{X_{j}}\) denoting the primary split
based on variable \emph{\(X_{i}\)} and the surrogate split on variable
\emph{\(X_{j}\)} of the node \emph{n} and \(nodes(X_{i})\) denoting the
total number of nodes based on \(X_{i}\). Related features can
subsequently be selected by a threshold adjusted by a user defined
factor \emph{t} (default for \emph{t} is 5).

\hypertarget{mututal-forest-impact-mfi}{%
\subsection{Mututal forest impact
(MFI)}\label{mututal-forest-impact-mfi}}

Inspired by Nembrini, König, and Wright (2018), we developed an unbiased
approach for the analysis of variable relations based on surrogate
variables. For this purpose, pseudo data Z, which is uninformative but
shares the structure of the original data X, is generated by the
permutation of the features across observations. Thus Z contains as many
permuted variables p as X. Subsequently, the mean adjusted agreement of
the features is determined for both, X and Z and the novel relation
parameter MFI is defined by

\[\widehat{MFI_{X_{i},X_{j}}} =  \widehat{M_{X_{i},X_{j}}} - \widehat{M_{Z_{i},Z_{j}}}\]

Just as for the mean adjusted agreement, a value of 1 for the MFI of two
features corresponds to an exact agreement of their impact on the random
forest model.

\hypertarget{mutual-impurity-reduction-mir}{%
\subsection{Mutual impurity reduction
(MIR)}\label{mutual-impurity-reduction-mir}}

Not only the relation parameter, but also the importances obtained by
SMD are biased. For this reason, we define the novel, unbiased
importance measure MIR that does not evaluate the features individually
but also considers the relations between them. MIR is the sum of the AIR
of the individual feature and the AIR of the other features multiplied
by the corresponding relation parameter MFI:

\[\widehat{MIR_{X_{i}}} =  \widehat{AIR_{X_{i}}} + \sum_{j}^{p} \widehat{MFI_{X_{i},X_{j}}} \cdot \widehat{AIR_{X_{j}}}\]

\hypertarget{testing-procedures}{%
\subsection{Testing procedures}\label{testing-procedures}}

Also inspired by Nembrini, König, and Wright (2018), statistical testing
procedure are performed to select relevant and related parameters and
the following null hypotheses are tested:

\[H_0: MFI_{X_{i},X_{j}} \leq 0\] \[H_0: MIR_{X_{i}} \leq 0\]

For this, the respective importance and relation values are tested
against a null distribution. This null distribution is obtained by zero
and negative values, which are mirrored to obtain the corresponding
positive values, as proposed by Janitza, Celik, and Boulesteix (2018).
However, for this approach to be validly applied, there must be a
sufficient number of negative values and thus a sufficient number of
features \emph{p}. Because of this, we developed alternative,
permutation-based approaches to obtain the null distributions for
important and related features. For the selection of related features in
MFI, the permuted relations \(\widehat{M_{Z_{i}}}\) are utilized. Since
they only contain zero and positive values, the null distribution is
completed similarly as in Janitza, Celik, and Boulesteix (2018): The
non-zero values are mirrored to obtain the respective negative values of
the distribution. For the selection of important features in MIR, the
permuted relations \(\widehat{M_{Z}}\) are multiplied by permuted values
of AIR. This process is repeated multiple times to determine enough
values to describe the null distribution sufficiently.

\hypertarget{simulation-studies}{%
\subsection{Simulation studies}\label{simulation-studies}}

\hypertarget{bias-of-importance-and-relation-measures}{%
\subsubsection{Bias of importance and relation
measures}\label{bias-of-importance-and-relation-measures}}

In order to study the bias of the SMD importance and relation analysis,
we simulated two simple null scenarios meaning that none of the
simulated variables is associated with the outcome. For both, the sample
size was set to 100 and the simulation was replicated 1000 times. As in
Nembrini, König, and Wright (2018), the outcomes for classification,
regression and survival analyses were generated from a binomial
distribution with a probability of 0.5, a standard normal distribution
and an exponentially distributed survival (\(\lambda\)~=~0.5) and
censoring time (\(\lambda\)~=~0.1), respectively. \newline AIR, SMD
importance and relation analysis, as well as the new approaches MFI and
MIR were applied to the simulated data utilizing RF with 100 trees, an
\emph{mtry} of 3 and a minimal node size of 1. Furthermore, for SMD, MFI
and MIR 3 surrogates \emph{s} were determined in each split. The
variables of the two scenarios were simulated as follows:

\hypertarget{null-scenario-a-increasing-number-of-expression-possibilities}{%
\paragraph{Null scenario A: Increasing number of expression
possibilities:}\label{null-scenario-a-increasing-number-of-expression-possibilities}}

Inspired by Nembrini, König, and Wright (2018), nine nominal variables
\(X_1\),\ldots{}\(X_{9}\) with 2, 3, 4, 5, 6, 7, 8, 10 and 20 categories
were generated from a uniform distribution and one continuous variable
\(X_{10}\) from a standard normal distribution.

\hypertarget{null-scenario-b-increasing-minor-allel-frequency}{%
\paragraph{Null scenario B: Increasing minor allel
frequency:}\label{null-scenario-b-increasing-minor-allel-frequency}}

As in Nembrini, König, and Wright (2018), ten variables
\(X_1\),\ldots{}\(X_{10}\) with minor allele frequencies of 0.05, 0.10,
0.15, 0.20, 0.25, 0.30, 0.35, 0.40, 0.45, 0.50 from a binomial
distribution were simulated.

\hypertarget{correlation-study}{%
\subsubsection{Correlation Study}\label{correlation-study}}

To evaluate the selection of variables in the presence of correlations,
a simulation study from Seifert, Gundlach, and Szymczak (2019) was
carried out. The quantitative outcome \(Y\) is dependent on 6 relevant
variables \(X_1\), \ldots, \(X_{6}\) that were, just like three
non-relevant, outcome-independent variables \(X_7\), \ldots, \(X_{9}\),
sampled from a standard normal distribution N(0,1):

\[Y = X_1 + X_2 + X_3 + X_4 + X_5 + X_6 + \epsilon\]

The noise \(\epsilon\) followed a N(0, 0.2) distribution. In addition,
10 correlated variables (denoted as c\(X_1\), c\(X_2\), c\(X_3\),
c\(X_7\), c\(X_8\), c\(X_9\)) were generated for each of \(X_1\),
\(X_2\), \(X_{3}\), \(X_7\), \(X_8\) and \(X_{9}\) utilizing the
simulateModule function of the R package WGCNA. (Langfelder and Horvath
2008) Strong correlations (0.9) were used for \(X_1\) and \(X_7\),
medium correlations (0.6) for \(X_2\) and \(X_8\), and low correlations
(0.3) for \(X_3\) and \(X_9\). Furthermore, to reach a total number of
\(p_1\) = 1000 variables, additional independent, non-correlated
variables (ncV) were simululated using a standard normal distribution.
We will denote the variables \(X_1\), \ldots, \(X_{6}\) as well as
c\(X_1\), c\(X_2\) and c\(X_3\) as relevant and the other variables as
non-relevant. For a graphical summary of the simulation please refer to
the supplementary information in Seifert, Gundlach, and Szymczak (2019).
\newline We simulated data for 100 individuals, generated 50 replicates
and applied the importance measures and feature selection approaches
AIR, SMD and MIR utilizing random forests with 1000 trees, an
\emph{mtry} of 177 (\(\widehat{=}\)~\(p_1^{(3/4)}\)), and a minimal node
size of 1. For AIR and MIR, the p-value thresholds 0.001, 0.01 and 0.05
and for SMD and MIR \emph{s} values of 5, 10, 20 and 100 were used.
\newline In order to compare type I error rates, we also simulated a
null scenario with 1000 independent predictor variables from a standard
normal distribution and an independent binary outcome. 50 replicates for
50 cases and 50 controls were simulated and AIR, MIR and SMD were
applied with the same parameters as above. Type I error rates were
subsequently estimated by the division of selected variables and the
total number of predictor variables.

\hypertarget{realistic-study}{%
\subsubsection{Realistic Study}\label{realistic-study}}

For the comparison of the feature selection approaches under realistic
correlation structures, gene expression data sets were simulated as in
Seifert, Gundlach, and Szymczak (2019) and Degenhardt, Seifert, and
Szymczak (2019) utilizing the R package Umpire (Zhang, Roebuck, and
Coombes 2012) and a realistic covariance matrix. The covariance matrix
was generated from an RNA-microarray dataset of breast cancer patients
with \(p_2\) = 12 592 genes obtained from The Cancer Genome Atlas
(Network CGA, 2012). A multivariate normal distribution for a 12
592-dimensional random vector with a mean vector of zeroes was applied
to obtain gene expression values for cases and controls. For the cases,
25 variables for each effect size from the set
\{\({-2,-1,-0.5,0.5,1,2}\)\} were randomly chosen and the means of the
variables were increased accordingly. The other variables and the
covariance matrix were the same for cases and controls. Two data sets
with 100 controls and 100 cases were simulated for each set of 150
causal variables and the whole process including random selection of
causal variables was repeated 50 times. \newline As in Seifert,
Gundlach, and Szymczak (2019), we estimated the evaluation criteria
stability, classification error, empirical power and false positive rate
(FPR) for the comparison of the performance of AIR, MIR and SMD. The
approaches were applied using random forests with 5000 trees, an mtry of
1188 (\(\widehat{=}\)~\(p_2^{(3/4)}\)), and a minimal node size of 1.
For AIR and MIR, a p-value threshold of 0.01 and for SMD and MIR s
values of 50, 100, 200 and 500 were used. \newline Stability was
determined by the Jaccard's index (He and Yu 2010), the ratio of the
length of the intersection and the length of the union of the two sets
of selected variables. The classification error was determined using
each of the two data sets of each replicate to select variables, on
which RF models with the same parameters as for the feature selection
were trained. The other data set was subsequently used as validation set
and the mean classification error was calculated for each pair. The
empirical power was determined separately for each absolute effect size
by the fraction of correct selections among all replicates. To determine
the FPR, the null variables were defined differently for each replicate
characterized by other causal variables. Since variables at least
moderately correlated to causal variables are of interest as well, only
non-causal and non-correlated variables (correlation coefficient
\textless{} 0.2) were defined as null variables. The FPR was calculated
by dividing the number of selected null variables by the total numbers
of null variables.

\hypertarget{results}{%
\section{Results}\label{results}}

\hypertarget{simulation-studies-1}{%
\subsection{Simulation studies}\label{simulation-studies-1}}

\hypertarget{bias-of-relation-measures}{%
\subsubsection{Bias of relation
measures}\label{bias-of-relation-measures}}

\begin{figure}
\includegraphics[width=1\linewidth]{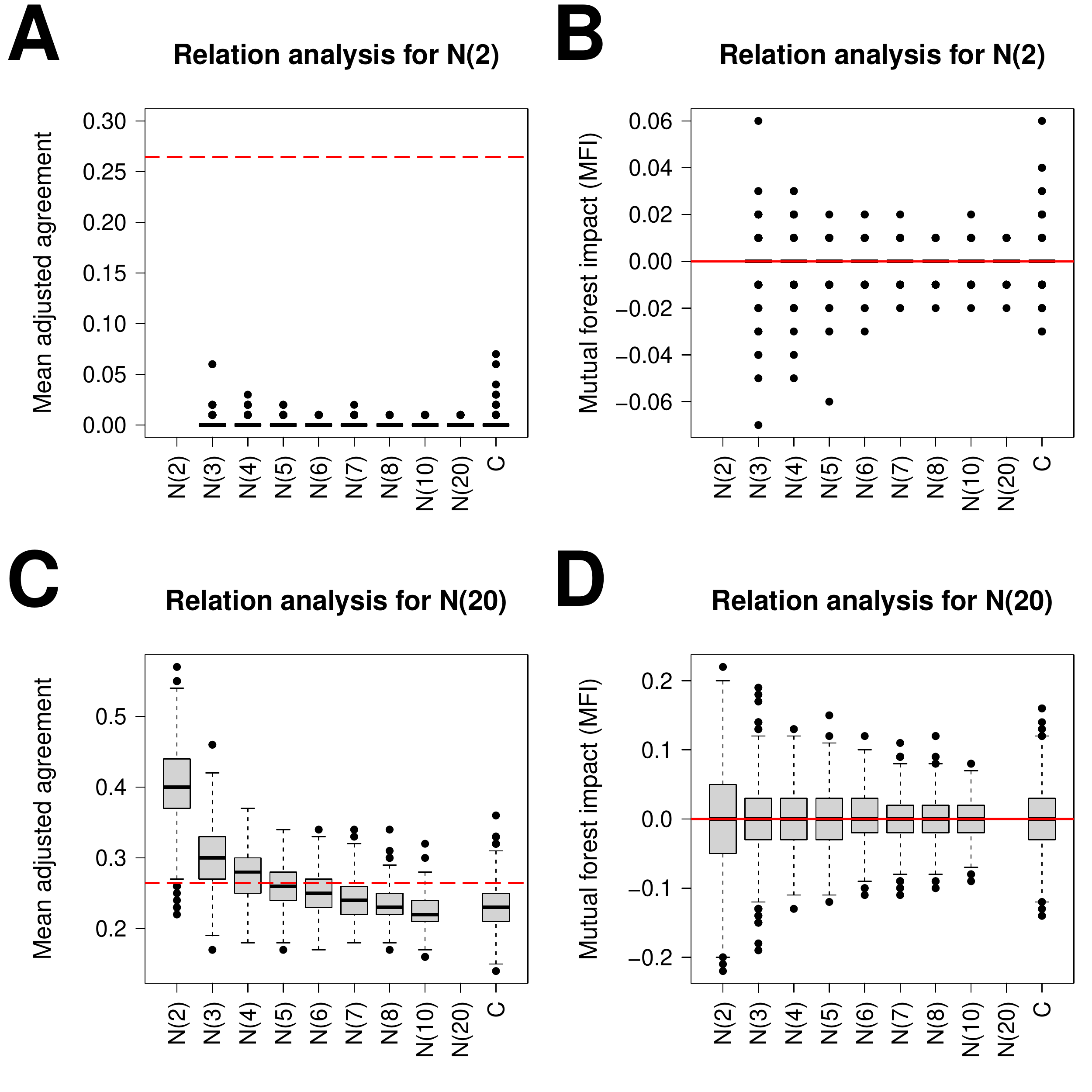} \caption{Null scenario A (classification): Relation analysis based on mean adjusted agreement (A+C) and MFI (B+D) for the nominal variables with 2 (A+B) and 20 (C+D) categories. The dashed line shows the threshold for the selection of related variables and a relation value of 0 for mean adjusted agreement (t=5) and MFI, respectively. The relations of the other variables are shown in the Supplementary Figures S1 and S2.}\label{fig:unnamed-chunk-1}
\end{figure}

The classification results for the null scenario with increasing number
of expression possibilities (null scenario A) are shown in Figure 1. The
mean adjusted agreement is 0 or very close to 0, when the relations of
the variable with 2 categories is analysed (Fig. 1A). Since all these
values are below the threshold, no variable is falsely selected here.
However, this does not apply to the relation analysis of the variable
with 20 categories (Fig. 1C). Because this variable has much more
expression possibilities, it shows increasing values for the mean
adjusted agreement as the number of categories decrease. For the
variable with only 2 categories, quite high values of approximately 0.4
are obtained resulting in a very frequent false selection of the
relation between these two variables. It is obvious that, similar as for
the importance analysis (Strobl et al. 2007), variables with many
categories are favoured in the relation analysis, especially when
relations to variables with low numbers of categories are analysed. MFI,
the novel approach for relation analysis, does not show this bias, since
all values for both, the variable with 2 and 20 categories, are located
around 0 (Fig. 1B+D). For the latter, however, the variance increases
for variables with lower numbers of categories. \newline Figure 2 shows
the results for the analysis of increasing MAF. The mean adjusted
agreement of variables with higher MAF is generally higher for the
relation analysis of both, the variable with a low MAF of 0.05 (Fig. 2A)
and the variable with a high MAF of 0.5 (Fig. 2C). Consequently,
relations of variables with high MAF are falsely selected more
frequently. The MFI is not influenced by the MAF in the same way,
because all variables show values around 0 for both variables (Fig.
2B+D). However, the variance increases for variables with higher MAF.
\newline The bias analysis of mean adjusted agreement and MFI, which is
shown here for a classification outcome, are also reflected in the
regression and survival analyses (see Supplementary Figures S5-S12).

\begin{figure}
\includegraphics[width=1\linewidth]{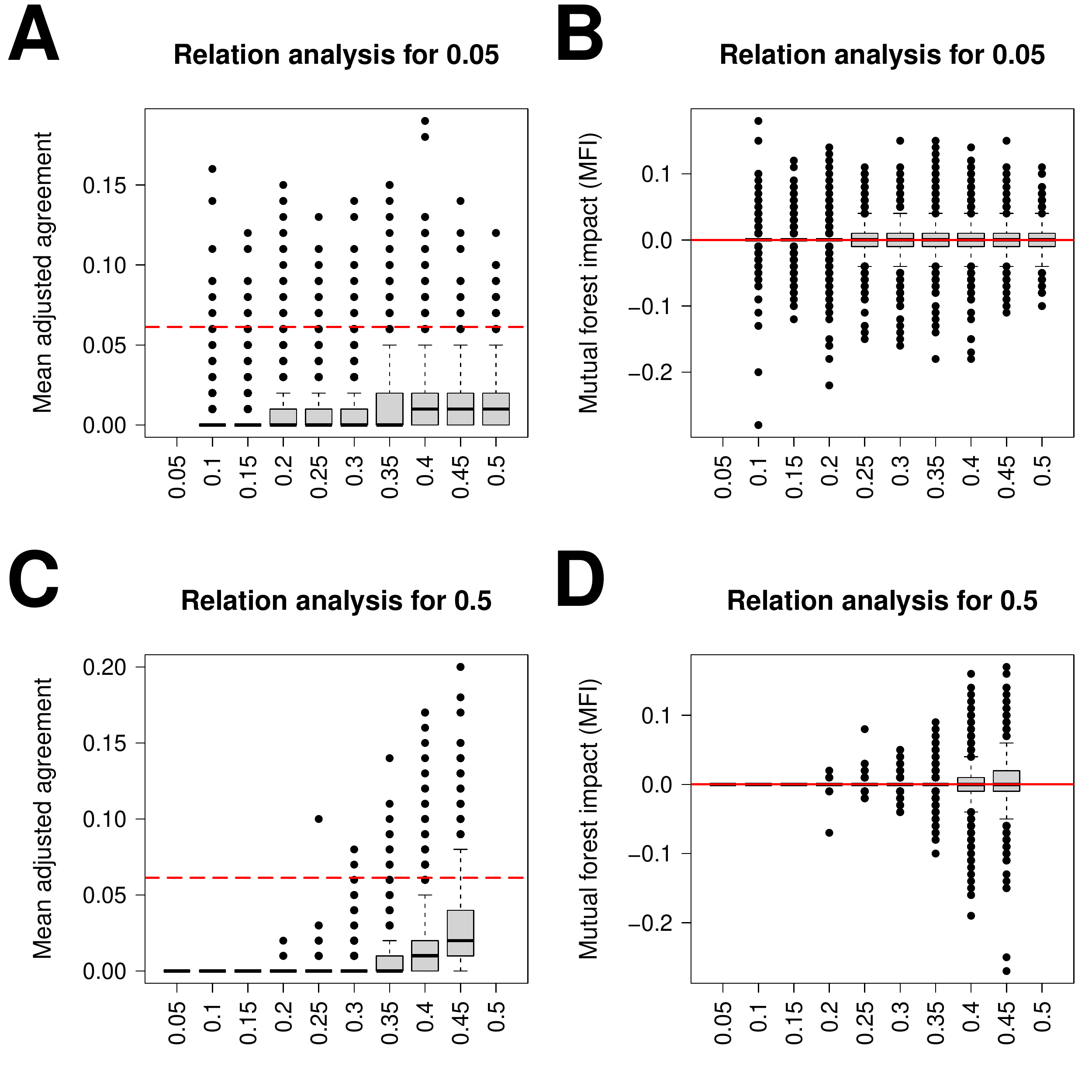} \caption{Null scenario B (classification): Relation analysis based on mean adjusted agreement (A+C) and MFI (B+D) for the variables with minor allele frequencies of 0.05 (A+B) and 0.5 (C+D). The dashed line shows the threshold for the selection of related variables and a relation value of 0 for mean adjusted agreement (t=5) and MFI, respectively. The relations of the other variables are shown in the Supplementary Figures S3 and S4.}\label{fig:unnamed-chunk-2}
\end{figure}

\hypertarget{bias-of-importance-measures}{%
\subsubsection{Bias of importance
measures}\label{bias-of-importance-measures}}

\begin{figure}
\includegraphics[width=0.8\linewidth]{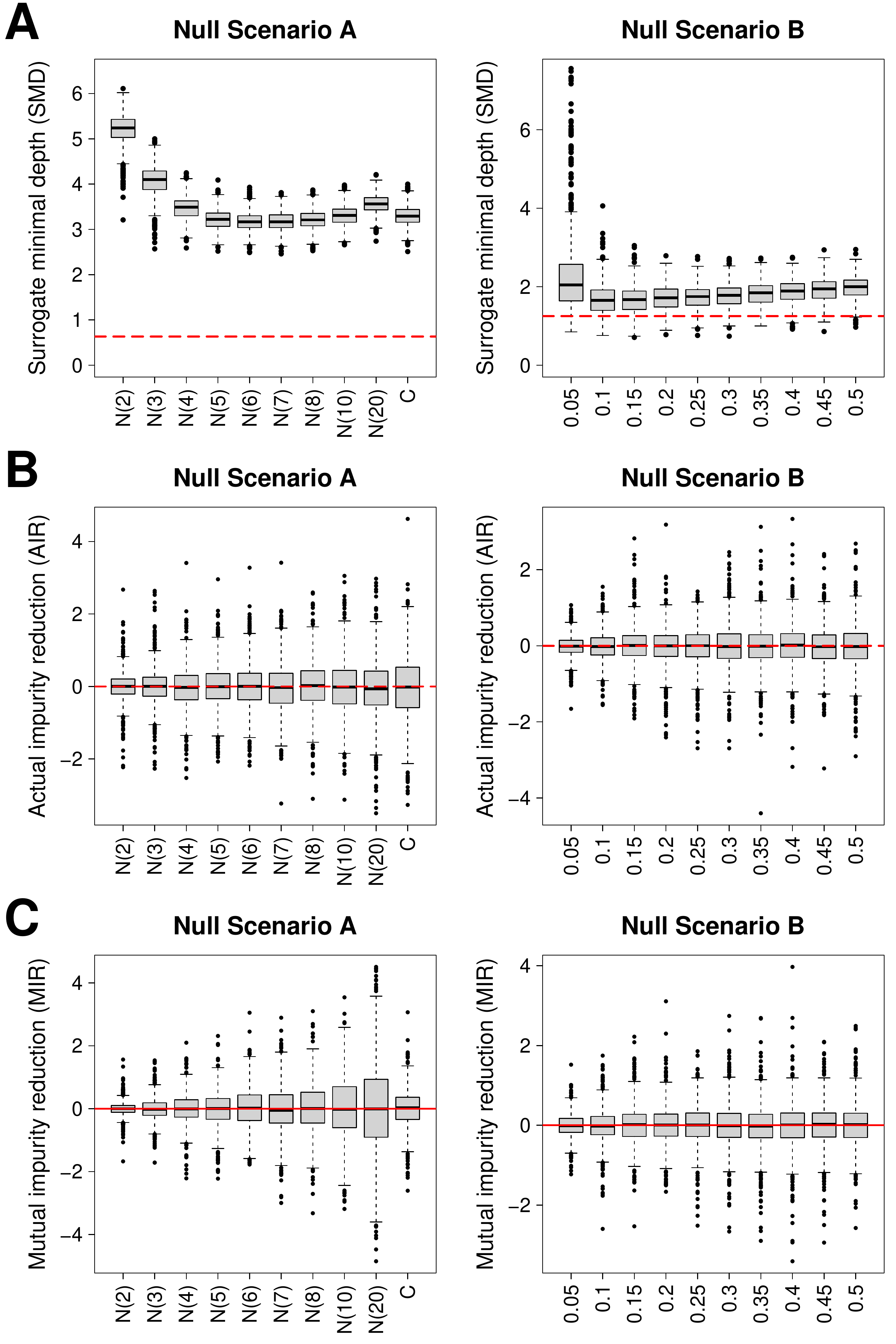} \caption{Variable importances for null scenarios (classification): Results of SMD (A), AIR (B), and MIR (C) in null scenario A (left) and B (right). The dashed line shows the threshold for the feature selection and an importance value of 0 for SMD (t=5) and AIR and MIR, respectively. Note that low SMD values correspond to important features. }\label{fig:unnamed-chunk-3}
\end{figure}

The results for the importance analyses for both null scenarios are
shown in Figure 3. SMD shows importance scores dependent on the number
of categories: (Fig. 1A left) Variables with low numbers of categories
have high values for SMD corresponding to low importances, while
variables with high numbers of categories seem more important because
they have lower SMD values. Hence, SMD shows the same bias as other
importance measures favouring variables with many categories (Strobl et
al. 2007). MIR, just like AIR, does not show this bias (Fig. 3B+C left).
However, the well-known property of higher variances for variables with
higher numbers of categories can be observed for AIR (Nembrini, König,
and Wright 2018). For MIR, this influence is even more evident. \newline
SMD is also influenced by the MAF showing different importance values
for variables with different MAF and consequently also different numbers
of false selections (Fig. 1A right). For AIR and MIR, this bias is not
observed. However, as expected from Nembrini, König, and Wright (2018),
the variance of the AIR increases towards higher MAF. For MIR, the same
property is apparent.

\hypertarget{correlation-study-1}{%
\subsubsection{Correlation Study}\label{correlation-study-1}}

\begin{figure}
\includegraphics[width=1\linewidth]{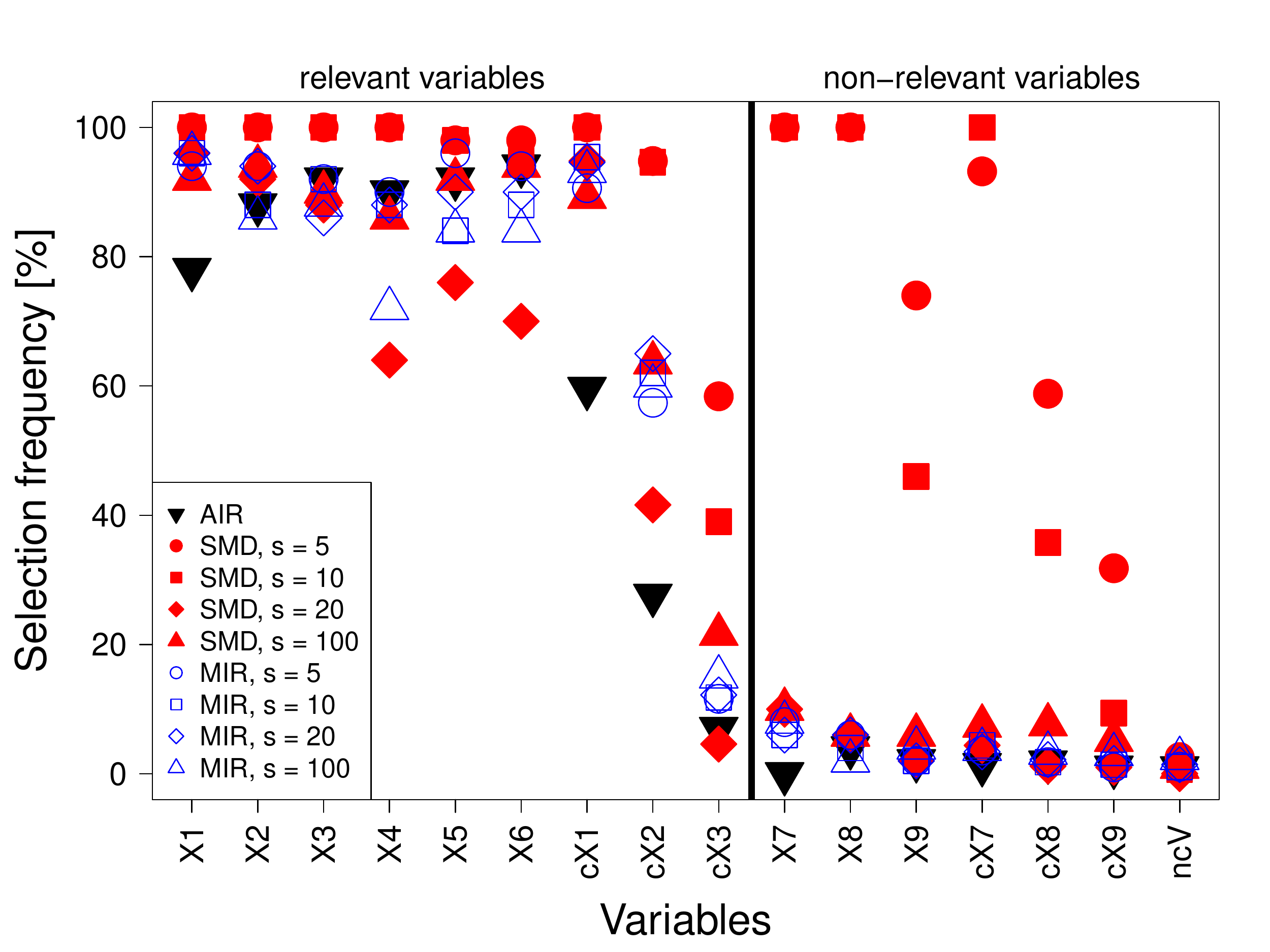} \caption{Results of the correlation study utilizing different numbers of surrogate variables (SMD and MIR) and a p-value thresholds of 0.01 (AIR and MIR). For the basic variables, ($X_1$ – $X_9$) the selection frequencies are averaged across all 50 replicates, whereas for the six groups of correlated variables, (c$X_1$ – c$X_3$ and c$X_7$ – c$X_9$) as well as the non-causal variables (ncV) the average frequencies across all replicates and variables in the respective group are shown.}\label{fig:unnamed-chunk-4}
\end{figure}

The selection frequencies of SMD, AIR and MIR for the different
variables and variable groups are shown in Figure 4. SMD with 5 and 10
surrogate variables (red circles and squares) shows the highest
selection frequencies for all relevant variables reaching 100\% for
\(X_1\), \(X_2\), \(X_3\), \(X_4\) and c\(X_1\), more than 90\% for
\(X_5\), \(X_6\) and c\(X_2\) as well as between 35 and 60\% for
c\(X_3\). However, also the correlated non-relevant variables are
frequently selected here, \(X_7\) and \(X_8\) even always. For SMD with
higher numbers of surrogate variables (red diamonds and triangles) and
for MIR, (blue symbols) the selection frequencies are mostly higher than
80\% for \(X_1\) to \(X_6\) and c\(X_1\). An exception are the
frequencies of SMD with s = 20 for \(X_4\) to \(X_6\) and MIR with s =
100 for \(X_4\) that are between 60 and 80\%. For c\(X_2\) and c\(X_3\),
MIR shows selection frequencies of around 60 and 15\% independent from
the number of surrogate variables used, while the frequencies of SMD are
much more dependent on this parameter. This is also evident for the
non-relevant variables, as MIR has similarly low frequencies, while SMD
is characterized by higher, more variable frequencies. \newline For AIR
(black triangle in Fig. 4), selection frequencies are lower than for MIR
and SMD (red and blue symbols) when correlations between the variables
exist. This applies for the causal variables, where the values for
\(X_1\), c\(X_1\) and c\(X_2\) are at around 80, 60 and 25\%,
respectively, and for the non-causal variables that show values at or
close to zero for all variables. \newline A comparison of different
p-value thresholds for MIR and AIR shows that the used value of 0.01,
the default value of AIR, is reasonable (Supplementary Figures S15). The
comparison of the selection of related variables by SMD and MIR shows no
significant differences, which is due to the fact that this study does
not use variables that exhibit the biases outlined above (Supplementary
Figures S16). \newline From the correlation study, it can be concluded
that MIR is more powerful for correlated variables than AIR and less
sensitive to changes in the number of surrogate variables used than SMD.
However, too high values for \emph{s} lead to an increased false
positive rate, which is apparent for the non-causal variables (ncV) in
Fig. 4 and the null scenario (Supplementary Figures S17).

\hypertarget{realistic-study-1}{%
\subsubsection{Realistic Study}\label{realistic-study-1}}

Figure 5 displays the results for the realistic simulation study. MIR
and SMD both show increasing empirical powers as the number of surrogate
variables \emph{s} increases (blue and red symbols in Fig. 5B). However,
for each value of \emph{s}, MIR has slightly higher empirical powers for
variables with low effect sizes (\textbar0.5\textbar). For the medium
effect size of \textbar1\textbar, this is only apparent for \emph{s} =
50 because empirical powers of 1 are achieved for higher values of
\emph{s} (blue and red circles in Fig. 5B). The stability shows an
opposite influence of this parameter, since high values are obtained
when \emph{s} is low and vice visa. When 100 and 200 surrogate variables
are used, SMD shows higher stabilities than MIR almost reaching 0.8 and
0.4, respectively (squares and diamonds in Fig. 5A). \newline AIR shows
similar empirical powers as SMD with \emph{s} = 100 (red square and
black triangle in Fig 5B). However the stability is much lower and, just
as in SMD and MIR with \emph{s} = 500 (red and blue triangles in Fig
5A), the false positive rate is higher than 0.

\begin{figure}
\includegraphics[width=1\linewidth]{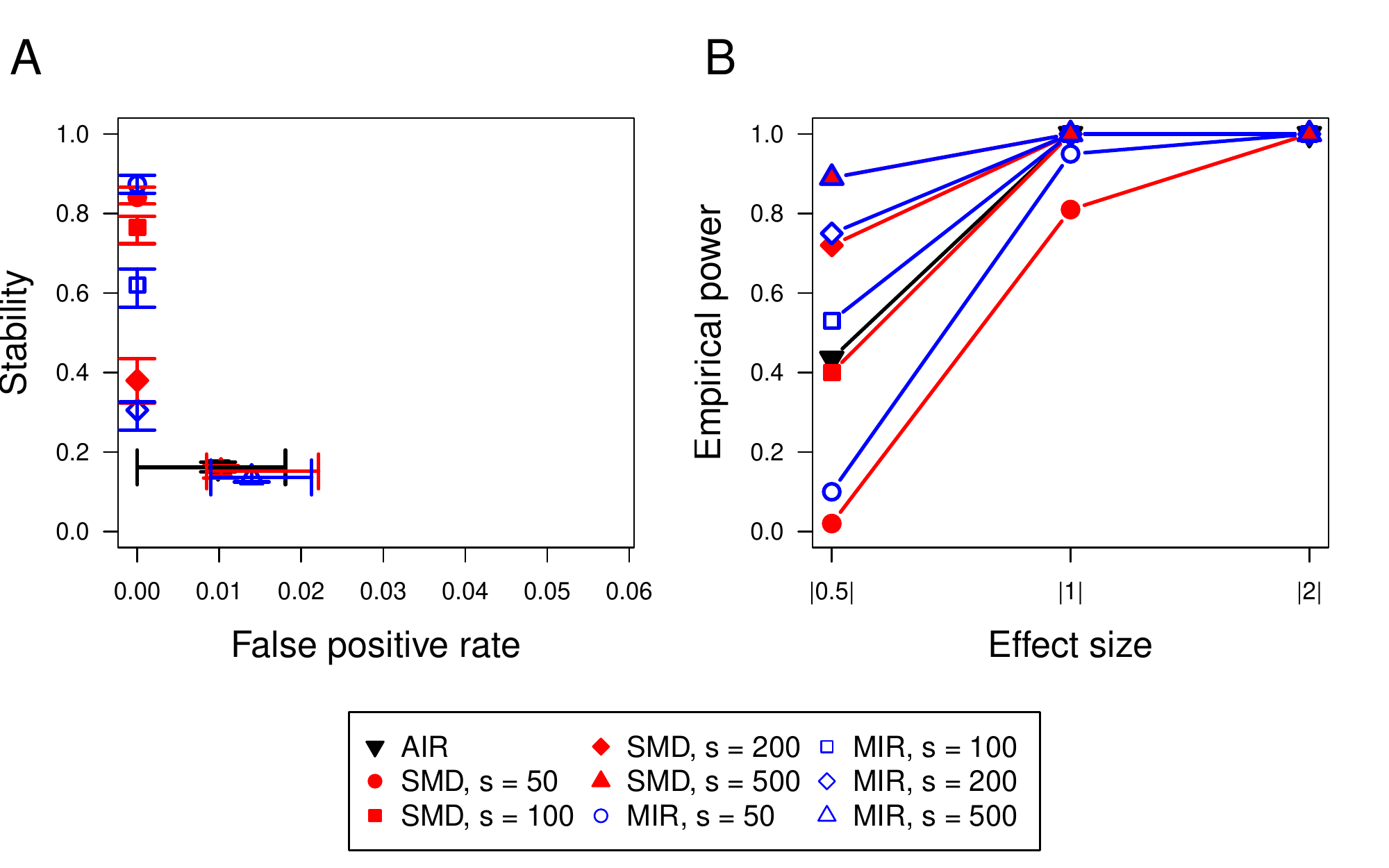} \caption{Results of the realistic study: Comparison of the performance of AIR, MIR (p-value threshold for both: 0.01) and SMD using different numbers of surrogates. Median, as well as the interquartile range over all 50 replicates for stability and false positive rate (A) and median frequencies of the empirical power for the different effect sizes (B) are shown. The classification error is 0 for all investigations.}\label{fig:unnamed-chunk-5}
\end{figure}

\hypertarget{discussion-and-conclusion}{%
\section{Discussion and conclusion}\label{discussion-and-conclusion}}

In this study, we introduced two novel approaches that are based on
surrogate variables in random forest: Mutual forest impact (MFI) for
relation analysis and mutual impurity reduction (MIR) as a feature
selection approach considering feature relations. We demonstrated that
both, MFI and MIR are not biased regarding variables with different
numbers of categories and category frequencies. They can be applied in
random forest classification, regression and survival analyses. \newline
Inspired by the feature selection based on the actual impurity reduction
(AIR) (Nembrini, König, and Wright 2018), we combined MFI and MIR with
the testing procedure of Janitza, Celik, and Boulesteix (2018) to
estimate p-values for the selection of related and important features,
respectively. The comparison of AIR and MIR demonstrated that the
inclusion of feature relations for selection is advantageous because a
higher stability and power is obtained. The comparison with SMD showed
lower probabilities of false selections and a smaller sensitivity to
changes of the crucial parameter \emph{s}, which determines the number
of surrogates used (Seifert, Gundlach, and Szymczak 2019). Since this
parameter is crucial to achieve an optimal compromise between high power
and low false selections, a smaller sensitivity to this parameter
enables the usage of higher values resulting in a higher power. However,
we have shown that even for MIR, too high values for \emph{s} should not
be used and additional analyses of data with different numbers of
features and correlation structures are needed to find optimal values
for the specific data types. We currently recommend for high-dimensional
settings to use 1 to 2\% of the total variables as utilized number of
surrogate variables in combination with a p-value threshold of 0.01.
\newline Due to the non-existing biases, a wide application of MFI and
MIR to data with different distributions, such as continuous
(e.g.~metabolomics, proteomics, isotopolomics), categorical (genomics)
or proportional data (epigenomics) is possible. MFI is especially
promising for the relation analysis of features across different data
sets, in which, in addition to different omics levels, even phenotypic
information and clinical biomarkers could be included. In addition omics
data could be combined with data from spectroscopic experiments, for
example based on infrared, Raman or X-ray radiation. \newline MFI allows
the selected features to be divided into groups with similar impact on
the random forest model. These groups and the feature relations they are
based on could be linked to known biological interactions, for example
in metabolic pathways, in a similar way as recently conducted with SMD
(Wenck et al. 2022). In subsequent analyses, the groups of related
features could be utilized instead of individual features to robustly
identify samples with specific properties, e.g.~by applying
pathway-based approaches (Seifert et al. 2020). In addition, based on
MFI, subgroups of samples could be revealed by the identification of
subgroup specific features. The identified subgroups could then be
applied for the characterization of the analysed samples and to improve
the machine learning model (Goel et al. 2020).\newline In conclusion,
the novel approaches MIR and MFI are very promising for the powerful
selection of relevant features and the comprehensive investigation of
the complex connection between features and outcome in omics,
multi-omics, and other data.

\hypertarget{acknowledgements}{%
\section*{Acknowledgements}\label{acknowledgements}}
\addcontentsline{toc}{section}{Acknowledgements}

This work is dedicated to Jakob Seifert. We want to thank Silke Szymczak
for constructive discussions. The results published here are in part
based on data generated by the Cancer Genome Atlas Research Network:
www.cancergenome.nih.gov.

\hypertarget{funding}{%
\section*{Funding}\label{funding}}
\addcontentsline{toc}{section}{Funding}

The research for this original paper was funded by the Deutsche
Forschungsgemeinschaft (DFG, German Research Foundation) under Germany's
Excellence Strategy -- EXC 2176 `Understanding Written Artefacts:
Material, Interaction and Transmission in Manuscript Cultures', project
no. 390893796. The research was conducted within the scope of the Centre
for the Study of Manuscript Cultures (CSMC) at Universität Hamburg.

\hypertarget{references}{%
\section*{References}\label{references}}
\addcontentsline{toc}{section}{References}

\hypertarget{refs}{}
\begin{CSLReferences}{1}{0}
\leavevmode\vadjust pre{\hypertarget{ref-boulesteix2012bib}{}}%
Boulesteix, A.-L., A. Bender, J. Lorenzo Bermejo, and C. Strobl. 2012.
{``Random Forest {Gini} Importance Favours {SNPs} with Large Minor
Allele Frequency: Impact, Sources and Recommendations.''} \emph{Brief.
Bioinform.} 13 (3): 292--304. \url{https://doi.org/10.1093/bib/bbr053}.

\leavevmode\vadjust pre{\hypertarget{ref-breiman2001ml}{}}%
Breiman, Leo. 2001. {``Random {Forests}.''} \emph{Mach. Learn.} 45 (1):
5--32. \url{https://doi.org/10.1023/A:1010933404324}.

\leavevmode\vadjust pre{\hypertarget{ref-breiman1984}{}}%
Breiman, Leo, Jerome Friedman, Charles J. Stone, and R. A. Olshen. 1984.
\emph{Classification and {Regression Trees}}. {Taylor \& Francis}.

\leavevmode\vadjust pre{\hypertarget{ref-chen2013ecaam}{}}%
Chen, Tianlu, Yu Cao, Yinan Zhang, Jiajian Liu, Yuqian Bao, Congrong
Wang, Weiping Jia, and Aihua Zhao. 2013. {``Random {Forest} in {Clinical
Metabolomics} for {Phenotypic Discrimination} and {Biomarker
Selection}.''} \emph{Evid. Based Complementary Altern. Med.} 2013:
1--11. \url{https://doi.org/10.1155/2013/298183}.

\leavevmode\vadjust pre{\hypertarget{ref-chen2012g}{}}%
Chen, Xi, and Hemant Ishwaran. 2012. {``Random Forests for Genomic Data
Analysis.''} \emph{Genomics} 99 (6): 323--29.
\url{https://doi.org/10.1016/j.ygeno.2012.04.003}.

\leavevmode\vadjust pre{\hypertarget{ref-debeer2020bb}{}}%
Debeer, Dries, and Carolin Strobl. 2020. {``Conditional Permutation
Importance Revisited.''} \emph{BMC Bioinform.} 21 (1): 307.
\url{https://doi.org/10.1186/s12859-020-03622-2}.

\leavevmode\vadjust pre{\hypertarget{ref-degenhardt2019bb}{}}%
Degenhardt, Frauke, Stephan Seifert, and Silke Szymczak. 2019.
{``Evaluation of Variable Selection Methods for Random Forests and Omics
Data Sets.''} \emph{Brief. Bioinform.} 20 (2): 492--503.
\url{https://doi.org/10.1093/bib/bbx124}.

\leavevmode\vadjust pre{\hypertarget{ref-goel2020a}{}}%
Goel, Karan, Albert Gu, Yixuan Li, and Christopher Ré. 2020. {``Model
{Patching}: {Closing} the {Subgroup Performance Gap} with {Data
Augmentation}.''} \emph{arXiv:2008.06775}, August.
\url{https://arxiv.org/abs/2008.06775}.

\leavevmode\vadjust pre{\hypertarget{ref-he2010cbac}{}}%
He, Zengyou, and Weichuan Yu. 2010. {``Stable Feature Selection for
Biomarker Discovery.''} \emph{Comput. Biol. Chem.} 34 (4): 215--25.
\url{https://doi.org/10.1016/j.compbiolchem.2010.07.002}.

\leavevmode\vadjust pre{\hypertarget{ref-ishwaran2015ml}{}}%
Ishwaran, Hemant. 2015. {``The Effect of Splitting on Random Forests.''}
\emph{Mach. Learn.} 99 (1): 75--118.
\url{https://doi.org/10.1007/s10994-014-5451-2}.

\leavevmode\vadjust pre{\hypertarget{ref-ishwaran2008aas}{}}%
Ishwaran, Hemant, Udaya B. Kogalur, Eugene H. Blackstone, and Michael S.
Lauer. 2008. {``Random Survival Forests.''} \emph{Ann. Appl. Stat.} 2
(3). \url{https://doi.org/10.1214/08-AOAS169}.

\leavevmode\vadjust pre{\hypertarget{ref-ishwaran2011sadm}{}}%
Ishwaran, Hemant, Udaya B. Kogalur, Xi Chen, and Andy J. Minn. 2011.
{``Random Survival Forests for High-Dimensional Data.''} \emph{Stat.
Anal. Data Min.} 4 (1): 115--32.
\url{https://doi.org/10.1002/sam.10103}.

\leavevmode\vadjust pre{\hypertarget{ref-ishwaran2010jasa}{}}%
Ishwaran, Hemant, Udaya B. Kogalur, Eiran Z. Gorodeski, Andy J. Minn,
and Michael S. Lauer. 2010. {``High-{Dimensional Variable Selection} for
{Survival Data}.''} \emph{J. Am. Stat. Assoc.} 105 (489): 205--17.
\url{https://doi.org/10.1198/jasa.2009.tm08622}.

\leavevmode\vadjust pre{\hypertarget{ref-janitza2018adac}{}}%
Janitza, Silke, Ender Celik, and Anne-Laure Boulesteix. 2018. {``A
Computationally Fast Variable Importance Test for Random Forests for
High-Dimensional Data.''} \emph{Adv. Data Anal. Classif.} 12: 885--915.

\leavevmode\vadjust pre{\hypertarget{ref-kursa2010jss}{}}%
Kursa, Miron B., and Witold R. Rudnicki. 2010. {``Feature {Selection}
with the {Boruta Package}.''} \emph{J. Stat. Softw.} 36 (1): 1--13.
\url{https://doi.org/10.18637/jss.v036.i11}.

\leavevmode\vadjust pre{\hypertarget{ref-langfelder2008bb}{}}%
Langfelder, Peter, and Steve Horvath. 2008. {``{WGCNA}: An {R} Package
for Weighted Correlation Network Analysis.''} \emph{BMC Bioinform.} 9:
559. \url{https://doi.org/10.1186/1471-2105-9-559}.

\leavevmode\vadjust pre{\hypertarget{ref-nembrini2018b}{}}%
Nembrini, Stefano, Inke R König, and Marvin N Wright. 2018. {``The
Revival of the {Gini} Importance?''} \emph{Bioinformatics} 34 (21):
3711--18. \url{https://doi.org/10.1093/bioinformatics/bty373}.

\leavevmode\vadjust pre{\hypertarget{ref-nicholls2020fg}{}}%
Nicholls, Hannah L., Christopher R. John, David S. Watson, Patricia B.
Munroe, Michael R. Barnes, and Claudia P. Cabrera. 2020. {``Reaching the
{End-Game} for {GWAS}: {Machine Learning Approaches} for the
{Prioritization} of {Complex Disease Loci}.''} \emph{Front. Genet.} 11
(April): 350. \url{https://doi.org/10.3389/fgene.2020.00350}.

\leavevmode\vadjust pre{\hypertarget{ref-nicodemus2011bb}{}}%
Nicodemus, K. K. 2011. {``Letter to the {Editor}: {On} the Stability and
Ranking of Predictors from Random Forest Variable Importance
Measures.''} \emph{Brief. Bioinform.} 12 (4): 369--73.
\url{https://doi.org/10.1093/bib/bbr016}.

\leavevmode\vadjust pre{\hypertarget{ref-nicodemus2010bb}{}}%
Nicodemus, Kristin K., James D. Malley, Carolin Strobl, and Andreas
Ziegler. 2010. {``The Behaviour of Random Forest Permutation-Based
Variable Importance Measures Under Predictor Correlation.''} \emph{BMC
Bioinform.} 11: 110. \url{https://doi.org/10.1186/1471-2105-11-110}.

\leavevmode\vadjust pre{\hypertarget{ref-sandri2008jocags}{}}%
Sandri, Marco, and Paola Zuccolotto. 2008. {``A {Bias Correction
Algorithm} for the {Gini Variable Importance Measure} in {Classification
Trees}.''} \emph{J. Comput. Graph. Stat.} 17 (3): 611--28.
\url{https://doi.org/10.1198/106186008X344522}.

\leavevmode\vadjust pre{\hypertarget{ref-seifert2020sr}{}}%
Seifert, Stephan. 2020. {``Application of Random Forest Based Approaches
to Surface-Enhanced {Raman} Scattering Data.''} \emph{Sci. Rep.} 10 (1):
5436. \url{https://doi.org/10.1038/s41598-020-62338-8}.

\leavevmode\vadjust pre{\hypertarget{ref-seifert2020b}{}}%
Seifert, Stephan, Sven Gundlach, Olaf Junge, and Silke Szymczak. 2020.
{``Integrating Biological Knowledge and Gene Expression Data Using
Pathway-Guided Random Forests: A Benchmarking Study.''} Edited by Luigi
Martelli. \emph{Bioinformatics} 36 (15): 4301--8.
\url{https://doi.org/10.1093/bioinformatics/btaa483}.

\leavevmode\vadjust pre{\hypertarget{ref-seifert2019b}{}}%
Seifert, Stephan, Sven Gundlach, and Silke Szymczak. 2019. {``Surrogate
Minimal Depth as an Importance Measure for Variables in Random
Forests.''} \emph{Bioinformatics} 35 (19): 3663--71.
\url{https://doi.org/10.1093/bioinformatics/btz149}.

\leavevmode\vadjust pre{\hypertarget{ref-shakiba2022mj}{}}%
Shakiba, Navid, Annika Gerdes, Nathalie Holz, Soeren Wenck, René
Bachmann, Tobias Schneider, Stephan Seifert, Markus Fischer, and Thomas
Hackl. 2022. {``Determination of the Geographical Origin of Hazelnuts
({Corylus} Avellana {L}.) By {Near-Infrared} Spectroscopy ({NIR}) and a
{Low-Level Fusion} with Nuclear Magnetic Resonance ({NMR}).''}
\emph{Microchem. J.} 174 (March): 107066.
\url{https://doi.org/10.1016/j.microc.2021.107066}.

\leavevmode\vadjust pre{\hypertarget{ref-strobl2008bb}{}}%
Strobl, Carolin, Anne-Laure Boulesteix, Thomas Kneib, Thomas Augustin,
and Achim Zeileis. 2008. {``Conditional Variable Importance for Random
Forests.''} \emph{BMC Bioinform.} 9: 307.
\url{https://doi.org/10.1186/1471-2105-9-307}.

\leavevmode\vadjust pre{\hypertarget{ref-strobl2007bb}{}}%
Strobl, Carolin, Anne-Laure Boulesteix, Achim Zeileis, and Torsten
Hothorn. 2007. {``Bias in Random Forest Variable Importance Measures:
{Illustrations}, Sources and a Solution.''} \emph{BMC Bioinform.} 8 (1):
25. \url{https://doi.org/10.1186/1471-2105-8-25}.

\leavevmode\vadjust pre{\hypertarget{ref-strobl2009pm}{}}%
Strobl, Carolin, James Malley, and Gerhard Tutz. 2009. {``An
Introduction to Recursive Partitioning: {Rationale}, Application, and
Characteristics of Classification and Regression Trees, Bagging, and
Random Forests.''} \emph{Psychol. Methods} 14 (4): 323--48.
\url{https://doi.org/10.1037/a0016973}.

\leavevmode\vadjust pre{\hypertarget{ref-szymczak2016bm}{}}%
Szymczak, Silke, Emily Holzinger, Abhijit Dasgupta, James D. Malley,
Anne M. Molloy, James L. Mills, Lawrence C. Brody, Dwight Stambolian,
and Joan E. Bailey-Wilson. 2016. {``{r2VIM}: {A} New Variable Selection
Method for Random Forests in Genome-Wide Association Studies.''}
\emph{BioData Min.} 9 (1).
\url{https://doi.org/10.1186/s13040-016-0087-3}.

\leavevmode\vadjust pre{\hypertarget{ref-wenck2022m}{}}%
Wenck, Soeren, Marina Creydt, Jule Hansen, Florian Gärber, Markus
Fischer, and Stephan Seifert. 2022. {``Opening the {Random Forest Black
Box} of the {Metabolome} by the {Application} of {Surrogate Minimal
Depth}.''} \emph{Metabolites} 12 (1): 5.
\url{https://doi.org/10.3390/metabo12010005}.

\leavevmode\vadjust pre{\hypertarget{ref-wright2017sm}{}}%
Wright, Marvin N., Theresa Dankowski, and Andreas Ziegler. 2017.
{``Unbiased Split Variable Selection for Random Survival Forests Using
Maximally Selected Rank Statistics.''} \emph{Stat. Med.} 36 (8):
1272--84. \url{https://doi.org/10.1002/sim.7212}.

\leavevmode\vadjust pre{\hypertarget{ref-zhang2012ijals}{}}%
Zhang, Jiexin, Paul L. Roebuck, and Kevin R. Coombes. 2012.
{``Simulating Gene Expression Data to Estimate Sample Size for Class and
Biomarker Discovery.''} \emph{Int. J. Advances Life Sci.} 4: 44--51.

\leavevmode\vadjust pre{\hypertarget{ref-zivanovic2019an}{}}%
Zivanovic, Vesna, Stephan Seifert, Daniela Drescher, Petra Schrade,
Stephan Werner, Peter Guttmann, Gergo Peter Szekeres, et al. 2019.
{``Optical {Nanosensing} of {Lipid Accumulation} Due to {Enzyme
Inhibition} in {Live Cells}.''} \emph{ACS Nano} 13 (8): 9363--75.
\url{https://doi.org/10.1021/acsnano.9b04001}.

\end{CSLReferences}

\bibliographystyle{unsrt}
\bibliography{bibliography.bib}

\end{document}